%% file: bare_jrnl.tex
\documentclass[journal]{IEEEtran}

\ifCLASSINFOpdf
\else
\fi

\usepackage{lineno,hyperref,amsmath,amsfonts}
\usepackage{color}
\usepackage{multirow}
\usepackage[ruled,vlined]{algorithm2e}

\usepackage{subfigure}
\usepackage{caption}
\usepackage{graphicx}
\usepackage[numbers,sort&compress]{natbib}
\newcommand{\venue}[1]{{\scriptsize #1}}
\newcommand{\bbest}[1]{{\bf #1}}

\newcommand{\best}[1]{\underline{#1}}

\begin{document}
\title{Feature Affinity based Pseudo Labeling for Semi-supervised Person Re-identification}
\author{Guodong Ding,
        Shanshan Zhang,
        Salman Khan,
        Zhenmin Tang,
        Jian Zhang, \IEEEmembership{Senior Member, IEEE} and
        Fatih Porikli, \IEEEmembership{Fellow, IEEE}
 \thanks{Guodong Ding, Shanshan Zhang and Zhenmin Tang are with the School of Computer Science and Engineering, Nanjing University of Science and Technology, Nanjing, Jiangsu, China. e-mails: guodong.ding@njust.edu.cn, shanshan.zhang@njust.edu.cn and tzm.cs@njust.edu.cn}
 \thanks{Salman Khan and Fatih Porikli are with College of Engineering and Computer Science, Australian National University, Australia. e-mails: salman.khan@anu.edu.au, fatih.porikli@anu.edu.au}
 \thanks{Jian Zhang is with School of Electrical and Data Engineering, University of Technology Sydney, Australia. e-mail: Jian.Zhang@uts.edu.au}
}

\markboth{Journal of \LaTeX\ Class Files,~Vol.~14, No.~8, August~2015}%
{Shell \MakeLowercase{\textit{et al.}}: Bare Demo of IEEEtran.cls for IEEE Journals}
\maketitle

\begin{abstract}
Person re-identification aims to match a person's identity across multiple camera streams. Deep neural networks have been successfully applied to the challenging person re-identification task. One remarkable bottleneck is that the existing deep models are data hungry and require large amounts of labeled training data. Acquiring manual annotations for pedestrian identity matchings in large-scale surveillance camera installations is a highly cumbersome task. 

Here, we propose the first semi-supervised approach that performs pseudo-labeling by considering complex relationships between unlabeled and labeled training samples in the feature space. Our approach first approximates the actual data manifold by learning a generative model via adversarial training. Given the trained model, data augmentation can be performed by generating new synthetic data samples which are unlabeled. An open research problem is how to effectively use this additional data for improved feature learning. To this end, this work proposes a novel Feature Affinity based Pseudo-Labeling (FAPL) approach with two possible label encodings under a unified setting. Our approach measures the affinity of unlabeled samples with the underlying clusters of labeled data samples using the intermediate feature representations from deep networks. FAPL trains with the joint supervision of cross-entropy loss together with a center regularization term, which not only ensures discriminative feature representation learning but also simultaneously predicts pseudo-labels for unlabeled data. Our extensive experiments on two standard large-scale datasets, Market-1501 and DukeMTMC-reID, demonstrate significant performance boosts over closely related competitors and outperforms state-of-the-art person re-identification techniques in most cases.
\end{abstract}

\IEEEpeerreviewmaketitle

\section{Introduction}
\label{sec:intro}
Person re-identification is one of the most popular but challenging research problem in computer vision. It has wide applications in video surveillance and human-computer interaction. Given a query image, the task is to search images from a large gallery that contain person with the same identity. The gallery images are usually captured from different cameras, viewpoints and at different time instances. Therefore, this problem setting introduces major challenges such as large variations in illumination, view-points, body postures and potential occlusions \cite{zheng2016person}. 

In recent years, deep Convolutional Neural Networks (CNNs) have achieved great success in person re-identification \cite{zheng2016person,huang2017deepdiff,qian2017multi,lin2017consistent}. However, CNN based methods are limited by the insufficient amount of data available for each identity. Manual labeling is the main bottleneck for acquiring large scale annotations due to the tedious and labor-intensive nature of the job. This problem is particularly more prominent for the case of person re-identification as the labeling involves manually selecting identities and associating images from different cameras with varying viewpoints, illumination, occlusions and body pose changes. This is evidenced by the fact that even recent large-scale datasets, such as Market-1501 and DukeMTMC-reID which have been acquired specifically for deep learning, still have very limited pedestrian images per identity. For example, Market-1501 dataset has on average 17.2 training images for 751 identities. Furthermore, the number of images is unevenly distributed such that some identities have as few as 2 samples, while only a few classes have more than 20 images. Therefore, it is highly important to perform intelligent data augmentation to extend the training set.

The emergence of Generative Adversarial Network (GAN) \cite{goodfellow2014generative} has partially addressed this  problem as they can generate novel images with good perceptual quality. However, a pressing issue is how to optimally use the synthetic unlabeled data as a regularizer to improve supervised learning. Initial efforts towards this problem adopted simplistic approaches e.g., \cite{odena2016semi} created a single new label for all generated images while \cite{salimans2016improved} used predictions from a pre-trained CNN model to label generated images. More recently, \cite{huang2018multi, Zheng2017Unlabeled} proposed to use Label Smooth Regularization (LSR) to assign pseudo-labels to synthetic data samples. LSR was proposed decades ago and revisited recently in \cite{szegedy2016rethinking} to reduce over-fitting by assigning small values to non-ground-truth classes (instead of 0) for cross-entropy loss computation. Specifically, \cite{Zheng2017Unlabeled} extends LSR to outliers (LSRO) by assigning uniformly distributed virtual labels to generated images from GAN networks. This choice was made to avoid emphatically classifying generated samples into one of the existing categories. Afterwards, \cite{huang2018multi} argued that generated images have considerable visual differences and assigning same labels to all would lead to ambiguous predictions. Thus, they proposed to provide labels based on the normalized class predictions (probability estimates) over all pre-defined classes.

One remarkable drawback of all existing pseudo labeling approaches is that they are agnostic to underlying relationships between an labeled and unlabeled data samples. The most mature effort in this direction \cite{huang2018multi} performs label generation based on class predictions that are solely dependent on the input sample and neglect distance based affinity between the unlabeled image and labeled examples. In this work, we attempt to overcome this shortcoming by dynamically associating unlabeled samples with pre-defined classes during the training process. Inspired by the spirit of clustering that leverages the underlying patterns within training data, we propose a novel label assigning approach called Feature Affinity based Pseudo-Labeling (FAPL) which delivers significant performance boost in person re-identification. FAPL aggregates labeled data samples that belong to the same class into refined clusters and simultaneously provides pseudo-labels to unlabeled data samples based on their similarity with each cluster center in the feature space. Building upon feature affinities, we propose two possible labeling schemes i.e., one-hot pseudo-label based on non-maximum suppression and distributed pseudo-label for soft label assignment. While the former is easier to implement and train, while the later one performs better in our evaluations.


Another observation that shed light on our work is that despite the one-hot and distributed label encodings derived from probability predictions have been proposed separately for unlabeled data, they fail to be combined together in a unified architecture. Previous efforts only considered unitary encoding, either one-hot or distributed. Specifically, \cite{odena2016semi,salimans2016improved,lee2013pseudo} chose one-hot labels for unlabeled data, while \cite{Zheng2017Unlabeled,huang2018multi} adopted distributed labels. The reason behind this situation is that the guarantee of a valid network training procedure is effective weight gradients, however, using probability predictions directly as distributed labels can not provide any weight corrections. \cite{huang2018multi} is a workaround proposing to assign labels based on probability rankings which is somehow effective but inescapably introduces errors. In this work, we endeavor to address this problem by introducing feature affinities. Feature affinity itself is not reverent to any class probability, granting its ability to produce pseudo-label with both encodings in a unified way while ensuring effective learning processes. 




Considering the recent progress in adversarial networks, we also study the effect of using better generative models for pseudo-labeling. Several efforts have recently been devoted to enhance the visual quality of synthetic images and stabilize the model training process \cite{goodfellow2014generative,mirza2014conditional,zhu2017unpaired,radford2015unsupervised,salimans2016improved,arjovsky2017wasserstein,gulrajani2017improved}. For this purpose, better loss functions have been explored as well as novel network configurations to generate realistic images. The use of extra information was investigated in Conditional GAN \cite{mirza2014conditional}, where both generator and discriminator are conditioned on extra information such as class labels, to improve the visual quality of generated samples. In this work, alongside DCGAN, we further experiment with the recent improved Wasserstein GAN (IWGAN) model which avoids modal collapse and generates high-quality samples with better convergence properties. The WGAN \cite{arjovsky2017wasserstein} is based on the Wasserstein distance measure as adversarial training loss function which is better suited to depict distances between distributions. Our experiments show that training with higher-quality images helps improve the re-identification performance.


The main contributions of this work are summarized as follows: 
\setlength{\itemsep}{0em}
\begin{itemize}

\item A multi-task loss formulation is proposed to for semi-supervised learning which have two advantages. First, it jointly considers inter-class and intra-class variations in feature space for more discriminative representation learning. Second, it can simultaneously estimate pseudo-labels for unlabeled data.

\item We first propose to consider feature affinities between GAN generated samples and labeled data rather than prediction probabilities to estimate pseudo-labels with two possible encodings. Besides, both encodings can be generated uniformly based on feature affinities.

\item Our experiments on two standard large-scale person re-identification benchmarks demonstrate that the proposed method achieves significant improvements over other pseudo-labeling approaches and also outperforms the best performing methods in most comparison cases.

\end{itemize}

The remainder of this paper is organized as follows. A review of related works is provided in Section \ref{sec:review}. In Section \ref{sec:approach}, we provide details of our proposed feature affinity based Pseudo-labeling approach followed by a discussion on why FAPL works better. Section \ref{sec:exp} exhibits the effectiveness of proposed methods on two standard person re-identification benchmarks and provides an extensive ablation study. We conclude with an outlook towards the future work in Section \ref{sec:conc}.

\section{Related work}
\label{sec:review}
In this section, we discuss the relevant works on semi-supervised learning, person re-identification with deep network architectures.

\subsection{Semi-supervised Learning}
Semi-supervised learning uses both labeled and unlabeled data to improve performance on a given task. It is driven by the its practical value in learning faster, cheaper, and better feature representations. In many real world applications, it is relatively easier to acquire a large amount of unlabeled data. Semi-supervised learning seeks to train a model that can make more accurate predictions on future unseen test data compared to a model learned only from labeled training data. Plenty of approaches have been proposed in the literature for this setting. Common semi-supervised learning methods include variants of generative models \cite{Kingma2014Semi}, graph Laplacian based methods \cite{Doquire2013A}, co-training \cite{Zhang2014Addressing}, and multi-view learning \cite{Zhu2016Multi}. Above works in semi-supervised learning are based on the fact that sufficient unlabeled data is available. However, collecting unlabeled data is also cumbersome in some applications. After the emergence of Generative Adversarial Network (GAN) \cite{goodfellow2014generative}, a branch of research on semi-supervised leaning has shifted to exploring GAN generated images \cite{odena2016semi, salimans2016improved}. This work, incorporates unlabeled samples generated by GAN alongside the real samples available in the labeled datasets.  

\subsection{Person re-identification}
Person re-identification aims at matching pedestrian images captured from different cameras that belong to the same identity. Main efforts in this area can be divided into two categories: (a) metric learning and (b) feature representation learning. Metric learning usually takes input in the form of image pairs or triplets and learn a similarity metric using pairwise or triplet loss \cite{cheng2016person,varior2016siamese,yi2014deep}. \cite{yi2014deep} propose to use deep networks to learn a similarity metric directly from image pixel, spare the trouble of feature crafting and engineering. \cite{cheng2016person} divides feature representations into four parts which are later concatenated for final triplet loss calculation. Metric learning has shown its great effectiveness in person re-identification, however, this stream of work suffers from huge data expansion when constituting image pairs and triplets especially when applied to large-scale datasets. \cite{Zhou2017Large} uses a part-based CNN to extract discriminative and stable feature representations and proposes a novel set-to-set (S2S) loss for similarity learning which ensures large margin between inter-class and intra-class set.

The other type of works focuses on feature learning, addressing this task in the form of classification. Common practices include first training  a pedestrian identity predicting model and then extract last fully connected layer activations as pedestrian descriptor for retrieval during testing \cite{zheng2016person,chen2017person,ding2017let,zheng2017discriminatively,varior2016siamese}. Amongst these, \cite{ding2017let} proposed to feed global info into previous layer for a compact feature representation, \cite{Zheng2017Unlabeled} demonstrated that the use of DCGAN generated samples to enlarge the training set helps achieve a boost in  performance. They proposed a label smoothing regularization for unlabeled samples (called LSRO) to assign a distributed pseudo-label. A major drawback of their approach is the underlying assumption that the synthetic data does not belong to any class, therefore considering a uniform distribution for all unlabeled samples. Our work aims to address this limitation and propose a novel loss function that automatically discovers patterns in the unlabeled data.

\section{Background on Pseudo-Labeling}
Pseudo-labeling is a technique to produce approximate labels for unlabeled data on the basis of labeled data instead of manually labeling them. As mentioned above, there have been works adopting GANs to generate samples in the field of person re-identification. Manually labeling those generated pedestrian images are even less practical as the image quality can not be assured would take huge effort to judge which identity it belongs to due to the variations in appearances. So pseudo-labeling for generated pedestrian images is vital for this case. Several works have been dedicated to find the optimal labeling scheme.

Existing pseudo-labeling approaches are depicted in Fig. \ref{fig:triplet} and summarized as follows: 
\begin{itemize}\setlength{\itemsep}{0em}
\item {\bf All-in-one} \cite{odena2016semi,salimans2016improved}. As illustrated in Fig. \ref{fig:allinone}, all-in-one seeks the easiest solution to assign labels. It simply introduces an extra new class and directly groups all unlabeled data into it without considering any variations which may exist between all generated images. Unlabeled data are trained with this fixed new class label throughout the training procedure.

\item {\bf One-hot} \cite{lee2013pseudo}. On the basis of all-in-one, One-hot takes into consideration the intra-variations within all generated image and propose to assume that each sample belongs to an existing class. Pseudo-labels are assigned by taking the maximum value for the probability prediction for each class shown as Fig. \ref{fig:onehot}. One-hot label is identical in its form to ground-truth label, thus this pseudo-labeling scheme is easy to train. Note that as the training proceeds, pseudo-labels for the same image can be different as the perditions might change. 

\input{tex/triplet}

\item {\bf Distributed} \cite{Zheng2017Unlabeled}, \cite{huang2018multi}. This type of pseudo-labeling further extends one-hot labeling scheme, and considers that the label for an unlabeled data should be distributed like $q_i$ in Fig. \ref{fig:distributed}. Since GAN generated images are fake samples drawn from the real data manifold, it would be inaccurate to classify them into any single class. Based on this assumption,  \cite{Zheng2017Unlabeled}  proposed to give equally distributed labels i.e., $q_i=1/K, \quad i=1,2,...,K$ in LSRO. In contrast, MpRL \cite{huang2018multi} assigns distributed labels according to class prediction ranks considering the class contributions. Distributed pseudo-labels together with real labels are trained with cross-entropy loss.
\end{itemize}

One proven ability of GAN is that it can generate samples from the training data distribution without strictly modeling it. In other words, GAN generated images can be seen as samples drawn from the labeled set. Thus, pseudo-labeling for generated images on the basis of labeled samples is more natural. However, none of existing pseudo-labeling methods takes into account the innate relations between labeled and unlabeled data to improve the feature representation learning under the semi-supervised framework. In contrast, this work aims to address this limitation and propose a novel loss function that automatically discovers patterns in the unlabeled data by associating them to labeled data samples. 
Next, we describe our semi-supervised learning approach based on pseudo-labeling.
 
\section{The proposed Approach}
\label{sec:approach}
\subsection{Overview}

\input{tex/archi}
Despite the fact that generated data samples from GAN are unlabeled, they can be used alongside the labeled examples to improve the learned features representations in a semi-supervised setting. We propose to take into account the underlying patterns in the labeled data and leverage those to infer pseudo-labels for unlabeled data. To this end we introduce a new multi-task learning objective for the semi-supervised training together with two new labeling schemes. The multi-task objective comprises of a normal classification loss and a center regularization term. The classification loss seeks to learn a ID-discriminative Embedding (IDE) for each pedestrian \cite{zheng2016mars}. The center regularization term improves the discriminative ability of feature embedding and simultaneously predict pseudo-labels for generated samples.

The overall network architecture is illustrated in Fig.~\ref{fig:archi}. The upper row denotes the training procedure of semi-supervised learning with synthetic images generated through GAN. It consists of two main modules. The \textbf{first} module on the left side is the image generation module, where a generative model is optimized using adversarial training to estimate the data distribution based on existing real samples (e.g., labeled images from training set). The generator is trained to create samples that pass the discriminator test, whilst the discriminator is trained to separate fake samples from real ones. Afterwards, the trained generator is be used to obtain large amounts of synthetic image samples lying on the approximated data manifold for subsequent training. The \textbf{second} module takes as input the generated data samples from which are unlabeled. This module uses the unlabeled data samples alongside the labeled ones to learn feature representations with the joint supervision of classification and a center regularization term.  The bottom row shows the testing phase, where  activations output from the convolutional neural network (CNN) are used as pedestrian descriptors for a Euclidean distance based retrieval operation.

The objective function (illustrated in Fig.~\ref{fig:archi}) of our proposed approach can be expressed as:
\begin{equation}
\mathcal{L} = \mathcal{L}_S+\lambda \mathcal{L}_C,
\label{eq:loss}
\end{equation}
where $\mathcal{L}_S$ and $\mathcal{L}_C$ respectively denote classification loss and center loss, and $\lambda$ is a trade-off parameter to balance the contribution of each component. We describe the motivation for the two loss terms in the following.



\subsection{Classification Loss}
Conventional supervised classification training requires {\it image-label} pairs, while labels are unavailable for generated data from a GAN model. In order to use the synthetic data for training, we propose two schemes to provide pseudo-labels for unlabeled data. Our empirical results show that both the approaches improve the person re-identification performance. We first provide notations and brief background and then elaborate on the approaches. 

For a single input image, the convolutional neural network calculates its feature representation $\mathbf{x}$ and output for $k$-th pre-defined class $y_k$, where $\mathbf{y} = \mathbf{W}^T \mathbf{x}+b$. It's estimated softmax probability to be classified into class $k$ can thus be given by:
\begin{equation}
p(y_k)=\frac{e^{y_k-y_{max}}}{\sum_{j=1}^K e^{y_j-y_{max}}}, \quad s.t.,\, k \in [1,K],
\end{equation}
where $y_{max}$ represents the maximum response in $\mathbf{y}$, $K$ is the number of pre-defined classes i.e., pedestrian identities in re-ID task.


\subsubsection{One-hot label}
One simple strategy is to assign a one-hot pseudo-label same as the real label following Fig. \ref{fig:onehot}. Referring to the clustering criterion and considering the similarity between GAN generated and real image representations, we propose a straightforward yet effective solution to associate an unlabeled data sample to the most similar class. 

The similarity between an input data representation in feature space $\mathbf{x}$ and cluster center $\mathbf{c}_k$ for class $k$ is formulated as below:
\begin{equation}
sim(\mathbf{x},\mathbf{c}_k) = \frac{\mathbf{x}\cdot \mathbf{c}_k}{\lVert \mathbf{x} \rVert \lVert \mathbf{c}_k \rVert}
\label{eq:sim}
\end{equation}
Pseudo-label $\ell$ for $\mathbf{x}$ is defined using the above mentioned similarity metric as follows:
\begin{equation}
\ell = \mathop{\arg\max}_{k}  \; sim(\mathbf{x},\mathbf{c}_k)
\label{eq:one}
\end{equation}

One advantage of one-hot pseudo-labeling is that it is consistent with ground truth labels, which enables unlabeled data to be integrated with labeled data for training without a separate training procedure with different loss formulation. They can be trained by following categorical loss function:
\begin{eqnarray}\label{eq:onefor}
\mathcal{L}_S&=&-\log(p(y_{\ell}))\\\nonumber
&=&-(y_{\ell} - y_{max}) +\log(\sum_{j=1}^K e^{y_j - y_{max}})
\end{eqnarray}
The backward gradients can be written as:
\begin{equation}\label{eq:onebac}
\mathcal{L}_S^{'} = \frac{e^{y_{\ell}-y_{max}}}{\sum_{j=1}^K e^{y_j - y_{max}}} - 1.
\end{equation}

\subsubsection{Distributed label}
The synthetic images generated by the GAN are random samples drawn from the approximated data manifold. Due to the complexity of high dimensional visual data, the pedestrian samples generated by GAN can have vague or absurd appearances and body shapes. These badly generated images succeed to pass the discriminator test yet they are easily distinguishable from the true samples when inspected by a human. 
Hence, it is inappropriate an optimal learning procedure to arbitrarily consider these images to belong to an existing identity and assign a one-hot label as discussed above. To this end, we propose to treat single unlabeled data approximately as a weighted combination of representations from different classes. This scheme is illustrated in Fig. \ref{fig:distributed} 

Accordingly, the final distributed label $q(y)$ is defined by a softmax function on the similarities between $\mathbf{x}$ and all cluster centers $c$, formulated as follows:
\begin{equation}
q(y_k) = \frac{e^{sim(\mathbf{x},\mathbf{c}_k)}}{\sum_{j=1}^K e^{sim(\mathbf{x},\mathbf{c}_j)}},
\label{eq:distributed}
\end{equation}
Distributed pseudo-label can thus be interpreted as the probabilities of unlabeled data belonging to each class. As suggested in \cite{szegedy2016rethinking}, cross-entropy function can be used to train with distributed pseudo-label, for a single input, its classification loss is calculated as:
\begin{eqnarray}\label{eq:disfor}
\mathcal{L}_S&=&-\sum_{k=1}^K q(y_k)\log(p(y_k))\\\nonumber
&=&-\sum_{k=1}^K q(y_k)(y_k-y_{max}) + q(y_k) \log(\sum_{j=1}^K e^{y_j-y_{max}})
\end{eqnarray}
The corresponding gradients are written as:
\begin{equation}
\mathcal{L}_S^{'} = q(y_k) \frac{e^{y_k-y_{max}}}{\sum_{j=1}^K e^{y_j - y_{max}}} - q(y_k)
\label{eq:disbac}
\end{equation}
Specifically, if we constrain $q(y_k)$  to satisfy:
\[q(y_k)=\begin{cases}
1&\text{$ k = \ell$},\\
0&\text{$k\neq \ell$}.
\end{cases}\]
and plug it in Eq. \ref{eq:disfor} and Eq. \ref{eq:disbac}, we can obtain Eq. \ref{eq:onefor} and Eq. \ref{eq:onebac}, respectively. This means distributed pseudo-labeling scheme is a generalization of one-hot pseudo-labels, and both encodings derived from feature affinity can be used for training in a unified architecture with the original cross-entropy loss function.

\subsection{Center Regularization} 
In addition to previous classification loss, we impose an extra center regularization term. This term associates each unlabeled sample to its matching cluster center in the feature space. It discovers underlying patterns in the feature space via center based clustering and thus performs intelligent data augmentation. We formulate the center regularization loss for a batch of $m$ image feature representations $\{\mathbf{x}_i \in \mathbb{R}^d\}_1^m$ as follows:
\begin{equation}
\mathcal{L}_C = \frac{1}{2} \sum_{i=1}^m \lVert \mathbf{x}_i - \mathbf{c}_{\ell_i} \rVert_2^2
\end{equation}
where $\mathbf{c}_{y_i} \in \mathbb{R}^d$ is the center with dimension $d$ of the cluster that $\mathbf{x}_i$ belongs to. This regularization term is applied to \textbf{(a)} prevent large intra-class variations which can not be addressed by classification loss alone, \textbf{(b)} obtain centers of all categories which in turn help decide the label for unlabeled data, whether one-hot or distributed. 

When losses are obtained, the backward gradients with respect to $\mathbf{x}_i$ can be calculated by:
\begin{equation}
\mathcal{L}'_C = \mathbf{x}_i - \mathbf{c}_{\ell_i}
\end{equation}
Next, the cluster centers are updated using the following equation:
\begin{equation}
\Delta \mathbf{c}_j = \frac{\sum_{i=1}^m  \delta (\ell_i=k)\cdot(\mathbf{c}_k-\mathbf{x}_i)}{1+\sum_{i=1}^m  \delta(\ell_i=k)}
\label{eq:center}
\end{equation}
where $\delta$ denotes delta function i.e., $\delta(condition)=1$ if $condition$ is satisfied, and otherwise 0.

We train network with three notable modifications: (a) Instead of taking the entire training set into account, centers are updated based on mini-batches. (b) Only labeled samples in mini-batches are allowed to update class centers which proves to help stabilize the training procedure since the generated may vary for a specific data in different training epoch, (c) The regularization loss is only back-propagated to labeled samples within each mini-batch.

The overall loss from Eq. \ref{eq:loss} for a batch of input samples for both one-hot and distributed pseudo-labeling approaches can be rewritten uniformly as follows:
\begin{equation}
\begin{aligned}
\mathcal{L} &= \mathcal{L}_S + \lambda \mathcal{L}_C\\
&=-\sum_{i=1}^m \sum_{k=1}^K q(y_{k}^i) \log (p(y_{k}^i)) + \frac{\lambda}{2} \sum_{i=1}^m \lVert \mathbf{x}_i - \mathbf{c}_{\ell_i} \rVert _2^2
\end{aligned}
\label{eq:finalloss}
\end{equation}
where $y_{k}^i$ represents $y_k$ for $i$-th input image.

The complete semi-supervised feature learning procedure is presented as Algorithm \ref{alg:overall}.

\input{tex/alg}

\subsection{Discussion}
In this section, we study the interesting properties of our proposed method and compare it with existing works. 

\subsubsection{Feature Similarity vs Class Predictions}
One significant difference between ours and all previous works on pseudo-labels generation is that this work is the first to propose assignment of pseudo-labels based on feature representation similarity in feature space. 

One common practice of deep re-id works is that they first train an identity classification network and then extract last fully connected layer activations as the final descriptor to perform similarity calculation during the subsequent testing phase. Previous works such as \cite{lee2013pseudo,huang2018multi} calculate pseudo-labels based on classification prediction probability. The network predictions can be directly used for one-hot pseudo-labeling \cite{lee2013pseudo} if the class with maximum probability response is used as a label for training. However, probability fails in the case of distributed labels since pseudo-labels would be identical to class probability predictions which will resultantly not produce any weight corrections based on back-propagated gradients. Therefore, \cite{huang2018multi} proposed to rank the predicted probabilities and assign labels based on the ranking, which inevitably introduces inaccuracies. On the contrary, we propose to regard pseudo-label generation itself as a retrieval process with unlabeled data as query and labeled data as gallery based on representation similarity, which is identical to the final retrieval performed in person re-identification. In this way, the similarity based labeling scheme can derive pseudo-labels in both one-hot and distributed cases.




\subsubsection{Why centers matter?} \label{par:why}
The contributions of center regularization term are two-fold: {\it 1) It promotes learning more discriminative feature representations for labeled data.} Conventional classification loss only considers classifying samples correctly, the resulting deeply learned features therefore contain large intra-class variations. \cite{wen2016discriminative} proposed a center loss jointly with the softmax loss to improve the discriminative power of the deeply learned features by reducing  intra-class variations. {\it 2) It produces pseudo-labels for unlabeled data considering their relationships with the labeled data.} This is an intuitive consideration because one can assume that the generated samples of a class are close to original ones.  In contrast, previous works such as \cite{lee2013pseudo} and \cite{huang2018multi} inappropriately label the data with predicted probabilities and do not take into account their inherent relationships with the labelled data. 

A toy example can be introduced to help illustrate above points. Left part in Fig. \ref{fig:center} shows a classification performed on a two-class (black and red) data set only with softmax loss. Unlabeled data samples (shown as blue triangles) are scattered around the boundary between two classes. Probability based labeling methods like \cite{lee2013pseudo} can be roughly seen as a nearest neighbor search. Sample 1 and 3 have as closest neighbor red points. Therefore, they are labeled red despite the fact in feature space they are more inclined towards black. Similar argument stands for data samples 2 and 4 being wrongly classified black. The right part is the labeling result after center loss is imposed. It is noticeable that less intra-class variations are introduced and unlabeled samples are more adequately classified considering the feature representation centers rather than the nearest neighbors.
\input{tex/center}

\subsubsection{Comparison with close work} The overall comparison of our approach with the closely related methods is summarized in Table \ref{tab:comp}. We denote our two schemes as FAPL-o (one-hot) and FAPL-d (distributed), respectively. Existing strategies for labeling GAN data in person re-identification include all-in-one \cite{odena2016semi,salimans2016improved}, one-hot \cite{lee2013pseudo}, LSRO \cite{Zheng2017Unlabeled} and dMpRL \cite{huang2018multi}. Their label distributions can be illustrated by Fig. \ref{fig:allinone}, Fig. \ref{fig:onehot} and Fig. \ref{fig:distributed}, respectively.  Both LSRO and dMpRL adopts distributed labels, the difference is that LSRO selects uniform distribution while dMpRL considers ranking contributions. 

 Compared with \cite{odena2016semi,Zheng2017Unlabeled} which directly assig fixed and identical labels for all generated data, our proposed considers the variations in between them and dynamically predicts pseudo-labels in each iteration as the training progresses. \cite{lee2013pseudo,huang2018multi} assign labels dynamically, both of which adopts class probability predictions rather than feature similarity to assign labels. We propose to take advantage of feature similarities in the feature space and predict  labels accordingly. Our experiments show that compared to the rigid {\it one-hot}  class labels, {\it distributed}  probability labels are more flexible and resilient. 

\input{tex/comp}

In summary, our proposed approach enjoys the benefits of being more flexible, discriminative and aware of wide context in the feature space. As a result, it leads to better performance as evidenced through the reported quantitative comparisons in Section \ref{sec:exp}.

\section{Experiments}
\label{sec:exp}
In this section, we perform experiments on two widely adopted large-scale person re-identification datasets to evaluate our proposed approach.

\subsection{Datasets and Evaluation Protocol.}
{\bf Market-1501} is a large-scale person re-identification dataset collected from 8 cameras on Tsinghua campus. In total it contains 12,936 images for training and 19,732 for testing, and the number of person identities for training and testing are 751 and 750, respectively. Overall, each identity in training set has 17.2 images on average. All the pedestrian images are automatically detected by Deformable Parts Model (DPM) \cite{felzenszwalb2010object}. 

{\bf DukeMTMC-reID} derives from a large multi-target, multi-camera pedestrian tracking dataset and released by \cite{Zheng2017Unlabeled}. Pedestrian images in this dataset are captured by 8 cameras with hand-labeled bounding boxes. It comprises of 1,404 identities in which 702 are selected as training set and the rest 702 are used for testing. The training set contains 16,522 images which leads to an average of 23.5 images per training identity. The query set has 2,228 images of 702 identities from one camera to retrieve from the gallery with 17,661 images.

{\bf Evaluation Metrics.} We evaluate our method with rank-1 accuracies and mean average precision (mAP) on Market-1501 and DukeMTMC-reID datasets. The rank-i accuracy denotes the rate at which one or more correctly matched images appear in top-i ranked images. The mAP value reflects the overall precision and recall rates, thus providing a more comprehensive evaluation metric.

\subsection{Implementation Details}

{\bf Re-id Baseline.} In our experiments, we adopt the standard ResNet-50 proposed in \cite{he2016deep} as the backbone architecture for our proposed approach. This network architecture has been used to evaluate closely related pseudo-labeling approaches, such as all-in-one\cite{salimans2016improved}, one-hot\cite{lee2013pseudo}, and LSRO\cite{Zheng2017Unlabeled}. No other changes were made to the architecture for training expect for substituting the last 1000 class activation neurons to target identity number, i.e., 751 and 702 for Market-1501 and DukeMTMC-reID, respectively. We first resize all training images to be $256 \times 256$ followed by a random horizontal flipping and cropping to the input size $224\times 224$. A dropout layer with 0.75 drop rate is inserted just before the final convolutional layer to prevent over-fitting for both datasets. The whole model is optimized by Stochastic Gradient Descent (SGD) with 0.9 momentum. The training is performed for a total of 50 epochs. During testing, last FC-layer with 2048-dim activations are extracted as the pedestrian descriptor for a cosine similarity based ranking. The network is implemented with the Matconvnet \cite{vedaldi2015matconvnet} package.

{\bf GAN Models.} For fair comparison, we follow \cite{Zheng2017Unlabeled} and adopt DCGAN \cite{radford2015unsupervised} as our model to generate fake unlabeled pedestrian images aiming at enlarging the training set. A 100-dim random is provided as the input to the \textbf{generator}, which is enlarged to form a 4$\times$4$\times$16 tensor by a linear function, followed by the application of 6 deconvolutional layers in total with $5 \times 5 $ sized kernels to obtain the desired 128$\times$128$\times$3 image. The \textbf{discriminator} has 5 convolutional layers with 5$\times$5 kernels to perform a binary classification task to separate real and fake samples from the given images. After the network is trained, we use the generator to produce up to 36,000 synthetic images. All synthetic data samples are resized to $256 \times 256$ for the following semi-supervised learning. Some generated data samples are displayed in Fig. \ref{fig:gans}, although some of the generated images far from the actual data distribution, they still help regularize the model and improve performance. We present our experimental results in Section \ref{subsec:eva}.

\subsection{Evaluation}
\label{subsec:eva}
\input{tex/effect}

\subsubsection{The effectiveness of proposed approach} Our overall results are summarized in Table \ref{tab:effect}. As shown, with the ResNet-50 as backbone architecture, the baseline achieved 72.74\%, 65.22\% rank-1 accuracy and 50.99\%, 44.99\% in mAP on Market-1501 and DukeMTMC-reID, respectively. We observe that the performance on DukeMTMC-reID is relatively lower than that on Market-1501, which is due to the heavy occlusions in the Duke dataset which makes the identification task more challenging. The goal here is to study the performance trend on the two datasets in comparison to other state of the art pseudo-labeling approaches. In this experiment, we randomly selected 24,000 generated images for two datasets as auxiliary data for training, and observed that both schemes lead to significant performance improvements over baseline. Rank-1 accuracy on Market-1501 increased by a margin of 9.30\% and 10.69\% for two schemes, respectively. mAP also enjoyed an increase from 50.99\% to 61.26\% and 63.23\%. Similar trend is observed on DukeMTMC-reID with an overall improvement of 6.20\% in rank-1 accuracy and 7.13\% in mAP.

\input{tex/gannumbers}

\subsubsection{Amount of unlabeled data} We use DCGAN to generate up to 36,000 images, from which we randomly pick subsets to evaluate how the total number of synthetic images incorporated for training influence the re-ID performance. Results are displayed in Table \ref{tab:gannumbers}. For our two labeling schemes, considerable performance increase in both rank-1 accuracy (9.38$\%$, 10.69\%) and mAP (11.32\%, 12.24\%) metrics was observed over baseline. However, an increase in the amount of unlabeled images above a threshold (i.e., from 12000 to 36000) failed to demonstrate considerable boost in performance and the final result fluctuates around 82$\%$ for one-hot and 83\% for distributed, respectively. Similar trend is observed amongst all other pseudo-labeling methods. We speculate that this phenomenon is due to the inherent representation ability of generated images. These images are sampled from a specific learned distribution (manifold of real data), therefore simply increasing sample numbers does not provide any additional information to the model that benefits the final retrieval task.

\input{tex/reduced}
\subsubsection{Amount of labeled data} It is desirable to learn better representation with less amount of labeled data. To test our approach when labeled data is extremely limited, we perform experiments on a reduced training set where available labeled examples are roughly cut to half and one third of the total amount. We follow the following rules when composing these subsets: (1) Keep all samples for identities with less than 8 images; 2) Keep half or  one-third samples form identities with more than 8 images and discard rest. We then obtain {\it half} subset with 7,106 training images and {\it one-third} subset with 4,200 training images. Results can be found in Table \ref{tab:reduced}. With labeled data reduced to a half and a third, performance dropped for the case of baseline model, from 72.74\% to 66.98\% and 57.45\% in rank-1 accuracy, respectively. This result is expected since less supervision is provided when training set is reduced. However, with our approach, we can observe an improvement around 10\% in rank-1 and 11\% in mAP over baseline for all rows. Specifically, with unlabeled data, the {\it half} model (rank-1= $\sim$77\%, mAP=55\%) managed to outperform the fully-supervised baseline (rank-1=72.74\%, mAP=50.99\%) by a significant margin of 5\%. Also note that the bottom line of subset {\it third}, unlabeled (36,000) images are roughly 8.5 times larger than labeled (4,200), the model still performed best (rank-1=67.87\%, mAP=43.54\%), which is a promising result showing that our approach can be applied on much smaller datasets.

\input{tex/param}
\subsubsection{Parameter sensitivity} We also conduct experiments studying the sensitivity of the trade-off parameter $\lambda$ in Eq. \ref{eq:finalloss}, whose results are presented in Table \ref{tab:param}. $\lambda$ is by default set to $10^{-4}$ in our experiments. We tried other settings for this parameter (such as $10^{-3}$ and $10^{-5}$) noticed a decrease in the performance of both one-hot and distributed labeling. For one-hot scheme, rank-1 accuracy dropped from 82.04\% to 80.82\% and 81.18\% and mAP from 61.26\% to 60.24\% and 59.42\%, respectively. Similar trend was observed on distributed labeling. Our empirical analysis showed that the choice of lambda used in our experiments roughly makes both loss terms comparable. 

\input{tex/gans}
\subsubsection{Unlabeled data with different visual quality} To better discover the effects that generated images with different quality has on regularizing the model, we select a state of the art GAN model to perform our evaluation. We choose the recently proposed  Improved Wasserstein GAN (IWGAN) \cite{gulrajani2017improved} for image generation. IWGAN has strong theoretical guarantees compared to DCGAN due to the use of Wasserstein distance measure as an adversarial training loss which provides faster and more stable convergence.

For the \textbf{generator}, we draw a 128-dim random noise vector and use five 3x3 residual (with skip connections) deconvolution layers to up-sample it to obtain 128x128x4 feature maps followed by another 3x3 convolution layer to generate the final 128x128x3 output sample. A \textbf{discriminator} takes as input 128x128x3 images and passes it first through a convolution layer to obtain a 128x128x64 intermediate presentation and then through another five residual down-sampling convolution layer to a 8192-dim representation followed by a binary classification similar to DCGAN to predict whether the input is real or fake. Similar to DCGAN image generation, output  images are resized to 256x256 for our model training.

We compare generated samples form both GAN networks in Fig. \ref{fig:gans}. One can notice that the generated images from both DCGAN and IWGAN are not comparable with real images, but it is clear that the IWGAN images shown in the bottom row are visually better than DCGAN images shown in the middle row. DCGAN  generated images have less diversity, contain considerable distortions as well as ambiguous limbs and body shapes. On the contrary, IWGAN better preserves human body shape and can generate more realistic samples with large color variations in clothing.

\input{tex/iwgan}

For each generation method, we randomly select varying numbers of fake images as an addition to our real training set and report results in Table \ref{tab:iwgan}. On Market-1501, under the same one-hot pseudo-labeling setting, DCGAN achieved 81.38\% in rank-1 while IWGAN achieved 83.28\%. In comparison, on DukeMTMC-reID, a maximum of 1.48\% rank-1 accuracy increase is observed when 12,000 samples are used. Overall, two conclusions can be drawn from this experiment:  (1) When better visual quality synthetic images are used from an improved GAN model, the performance across both datasets is further boosted by a margin of 0.5\%-1\%. This improvement is relatively small because sufficient samples from both models are considered which reduces the impact of bad quality samples. (2) Distributed pseudo-labeling approach consistently outperforms the one-hot approach no matter which GAN model is adopted.

\subsubsection{Comparison with other pseudo-labeling approaches} In this section, we compare the proposed approach with all four existing pseudo-labeling methods that we are aware of on Market-1501 dataset. The compared pseudo-labeling methods include all-in-one\cite{odena2016semi,salimans2016improved}, one-hot\cite{lee2013pseudo}, LSRO\cite{Zheng2017Unlabeled} and MpRL\cite{huang2018multi}. 

Among all published works, LSRO\cite{Zheng2017Unlabeled} is the state-of-art proposing to assign uniformly distributed pseudo-labels for unlabeled data to regularize the model. While MpRL\cite{Zhu2016Multi} is a very recent work by improving the distribution by considering class contributions and achieves very competitive results. Three different implementations, sMpRL, dMpRL-I and dMpRL-II, are provided in \cite{huang2018multi}. More specifically, the first sMpRL assigned fixed distributed labels throughout the whole training process, this is similar to LSRO except for class contributions are considered when labels are produced. Both dMpRL-I and dMpRL-II dynamically assign pseudo-labels to each generated samples, but differ in when unlabeled data are used for training. Generated data are used from the start point of training in dMpRL-I, while used after 20 epochs in dMpRL-II when the CNN network is relatively stable. 

Table \ref{tab:gannumbers} also summarizes the results of the state-of-the-art pseudo-labeling on person re-identification. It is shown that LSRO\cite{Zheng2017Unlabeled} achieved best performance with rank-1=78.21\% and mAP=56.33\% on Market-1501 dataset with 24,000 generated data. An overall performance increase amongst three MpRL\cite{huang2018multi} implementations can be observed, with dMpRL-II achieving the best results rank-1=80,37\% and mAP=58.59\%. Our proposed one-hot labeling scheme outperform best competitor dMpRL-II by a margin of 1.75$\%$ and 3.72$\%$ in rank-1 accuracy and mAP, respectively, and distributed scheme further improves the performance to 3.06\% and 4.64\%. This is reasonable since distributed labels consider similarity contributions from each class and are more suitable for GAN generated data. Another remarkable fact is that despite our proposed labeling strategies assign pseudo-labels to unlabeled data immediately as training starts, they both outperforms the dMpRL-II which only starts to produce labels with relative stable CNN network after several epochs. This comparison proves that our proposed methods is superior to all state-of-art pseudo-labeling methods.

\input{tex/market}
\input{tex/duke}
\subsubsection{Comparison with state-of-art methods} Our work is dedicated to better exploiting synthetic images to boost re-id performance rather than beating the state-of-the-art results, however, we still compare our proposed approach with state-of-the-art works on Market-1501 and DukeMTMC-reID datasets to show its competence. Our distributed labels with IWGAN images achieved rank-1=83.58\%, mAP=63.78\% on Market-1501 dataset, which is very competitive with many state-of-the-art methods except for JLML (rank1=85.1\%, mAP=65.50\%), PDC. The main reason why JLML outperforms by a margin of 2\% is because JLML adopted a much stronger baseline (around $3\%$ higher than ours) and incorporates three extra networks focusing on different local areas compared to our single branch architecture. With a state-of-art re-ranking technique from \cite{zhong2017re}, we observed a further boost of 2.5\% in rank-1 and 13.86\% mAP demonstrating that reciprocal relationships are encoded in our learned identity representations. On DukeMTMC-reID dataset, we achieved 79.04\% rank-1 accuracy and 70.74\% with re-ranking. DFPL slightly outperforms ours in rank-1 (around 0.2\%) because it takes advantage of multiple networks with different input scales and imposes consensus learning to force representations from different scales to be close if they belong to the same identity, while ours only adopts a single network. However, our mAP achieved 70.74\% which is 10\% higher than DFPL (60.60\%).

\subsection{Ablation Study}
We provide ablative experimental results on Market-1501 to evaluate each component of our proposed approach. The network is under full supervision of labeled data for baseline and center-loss and turns into a semi-supervised case when unlabeled data is introduced.

\begin{table}[h]
\caption{Ablative experiments in terms of each component of our proposed approach on Market-1501. 24,000 unlabeled data are incorporated for semi-supervised learning.}
\label{tab:ablative}
\centering
\begin{tabular}{c|cc|c}
\hline
Methods & rank-1 & mAP & supervision\\ \hline
Baseline & 72.74 & 50.99 & full\\
Center & 79.45 & 57.25 & full \\
FAPL-o & 82.04 & 61.26 & semi\\
FAPL-d & \bbest{83.43} & \bbest{63.23} & semi\\
\hline
\end{tabular}
\end{table}

\subsubsection{Center loss} \label{par:center} Center loss plays a crucial role in our proposed pseudo-labeling approach by reducing intra-class variations between data points, thus leading to more discriminative feature representations \cite{wen2016discriminative}. In this experiment, only labeled data is used for training to show the effectiveness of center regularization. When the center loss was applied (second row in Table. \ref{tab:ablative}), the baseline experienced a 6.71\% rank-1 increase from 72.74\% to 79.45\%  and 6.26\% gain in mAP from 50.99\% to achieve 57.25\%. This confirms the positive affect center loss has on learning more discriminative representations.

\subsubsection{Pseudo-labeling} On top of center loss, we add in our pseudo-labeling with both proposed schemes, each denoted as "FAPL-o" and "FAPL-d" in Table \ref{tab:ablative}. It can be observed that when synthetic data is incorporated for training with pseudo-labels, the network gains a further performance boost on both metrics to achieve 82.04\% rank-1 accuracy for one-hot and 83.43\% for distributed, 61.26\% in mAP for one-hot and 63.23\% for distributed, respectively.

\section{Conclusion}
\label{sec:conc}
In this paper, we emphasize on the fact that a reasonable labeling approach for GAN generated images should consider representation similarity and encode their relationships with real data samples. To this end, we proposed a Feature Affinity based Pseudo Labeling (FAPL) approach with one-hot and distributed label encodings for the person re-identification task. Unlabeled images are assigned label encodings according to their distance to identity centers in feature space and help address re-id problem in a semi-supervised manner. Experiment results show that our proposed approach outperforms other pseudo-labeling methods on person re-identification task by a large margin and achieves competitive accuracy compared to state-of-the-art solutions.

\ifCLASSOPTIONcaptionsoff
  \newpage
\fi



%

\bibliographystyle{IEEEtran}  
\bibliography{./bibtex/bib/myref} 
%








\end{document}

%% file: tex/triplet.tex
\begin{figure}[t]
\centering
\scalebox{0.5}{
\subfigure[All-in-one]{
\centering
\begin{minipage}{0.3\textwidth}
\includegraphics[width=\textwidth]{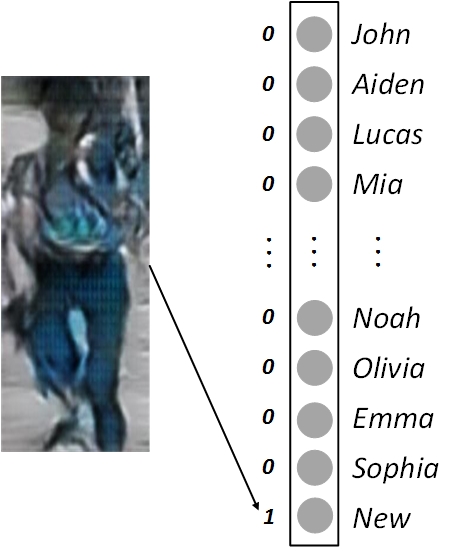}
\label{fig:allinone}
\end{minipage}
}
\subfigure[One-hot]{
\centering
\begin{minipage}{0.3\textwidth}
\includegraphics[width=\textwidth]{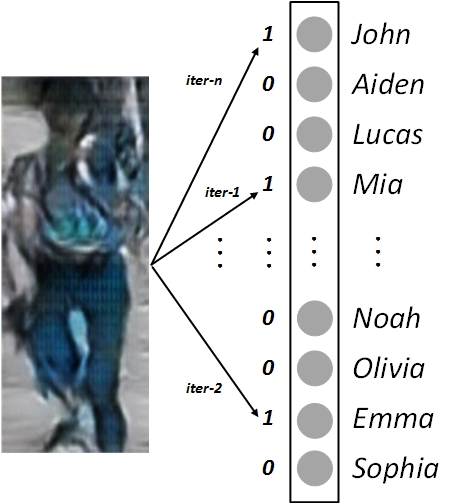}
\label{fig:onehot}
\end{minipage}
}
\subfigure[Distributed]{
\centering
\begin{minipage}{0.3\textwidth}
\includegraphics[width=\textwidth]{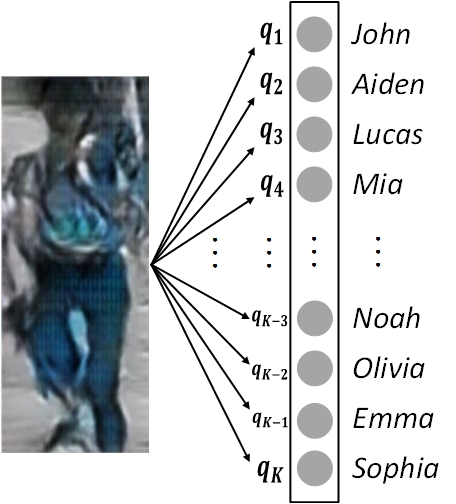}
\label{fig:distributed}
\end{minipage}
}
}
\caption{Existing pseudo-labeling approaches can be divided into above three categories. From left to right are: All-in-one, One-hot and Distributed. All-in-one treat all unlabeled data belong to a new class. One-hot assigns each unlabeled data a dynamic class label in each training epoch. Distributed considers contributions while labeling unlabeled data.}
\label{fig:triplet}
\end{figure}

%% file: tex/archi.tex

\begin{figure*}
\centering
\noindent\begin{minipage}[t]{.48\textwidth}
  \centering
  \includegraphics[width=0.4\textheight]{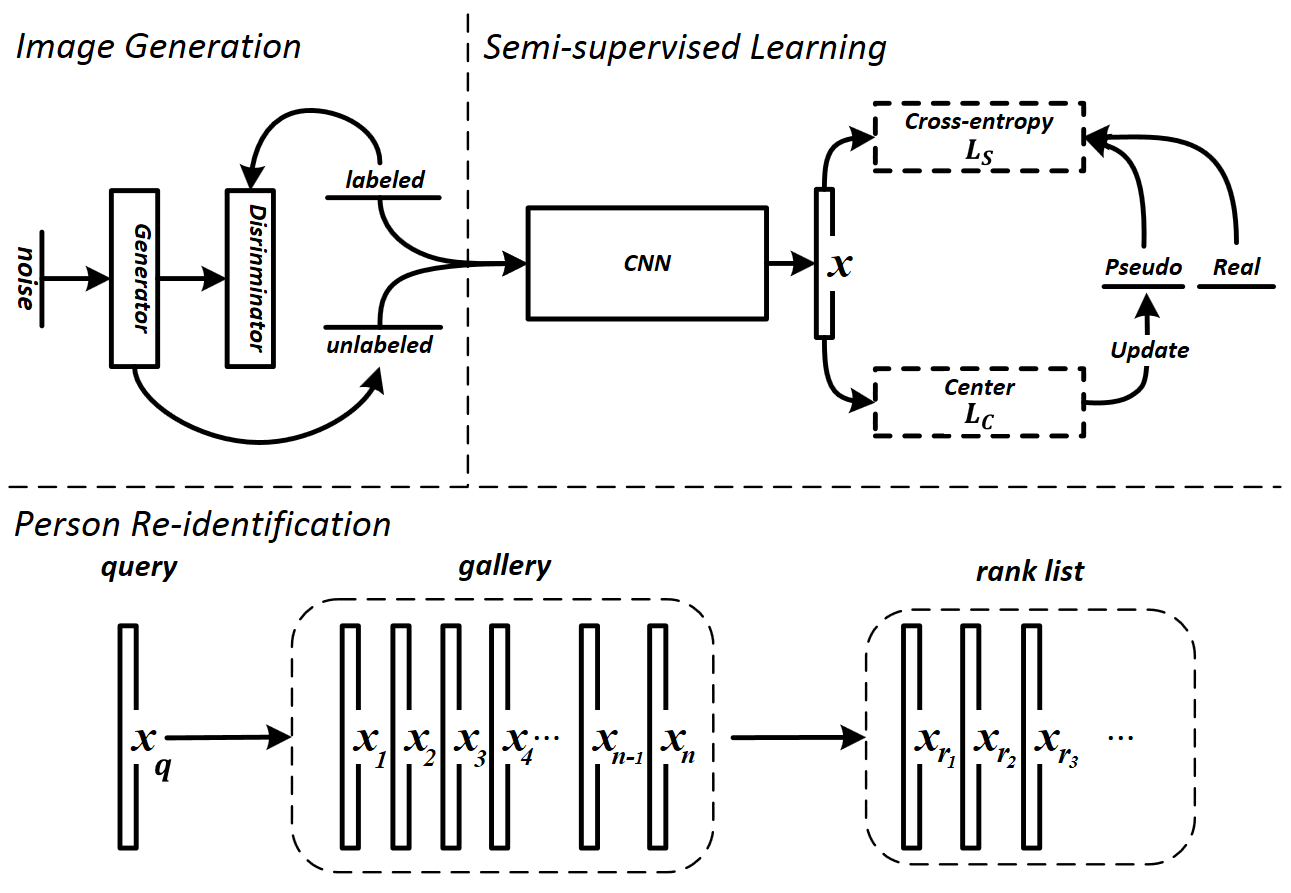}
\end{minipage}\hfill
\begin{minipage}[b]{.45\textwidth}
\captionof{figure}{The overall workflow of semi-supervised person re-identification. The top half denotes the training procedure while the bottom denotes the testing phrase. Left part of training denoted as image generation consists of a GAN network that has one generator network and one discriminator network. The former takes in as input some random noise and output a generated sample. The discriminator tries to distinguish generated from real samples. The right half demonstrates the semi-supervised training on a CNN architecture with our proposed multi-task loss (dashed boxes). Center loss performs clustering of unlabeled data in feature space and simultaneously outputs pseudo-labels for them. During testing, we extract CNN output $x$ as the pedestrian descriptor for both query and gallery images. Similar images are retrieved and ranked according to their descriptor similarity.}
\label{fig:archi}
\end{minipage}
\end{figure*}

%% file: tex/alg.tex
\begin{algorithm}[]
\KwIn{labeled data: $L$, unlabeled generated data: $U$, maximum iteration: $T$, batch size: $m$, center update rate: $\alpha$, trader-off parameter: $\lambda$, network parameters: $\theta$}
\KwOut{Optimized parameters $\hat{\theta}$} 
{\bf Initialization:} Training set $X = L \cup U$, Initialize $\theta$ with pre-trained ResNet-50, $\alpha=0.5$, $\lambda=10^{-4}$, class centers $\{\mathbf{c}_k = {\bf 0} | k=1,2,...,K \} $  
	
\nl \For{$t = 1:T$}{
Shuffle $X$ and sample $m$ samples to form a mini batch $X^t$\;
 Feed forward $X^t$ through CNN to obtain their feature representations $\mathbf{x}^t$\; 
    \For {$\mathbf{x}_i^t \in U$}{
    Calculate $sim(\mathbf{x}_i^t,\mathbf{c}_k)$ using Eq. \eqref{eq:sim}\;
    Generate pseudo label $\ell_{\mathbf{x}_i^t}$ for $\mathbf{x}_i^t$, one-hot with Eq. \ref{eq:one} or distributed with Eq.\ref{eq:distributed}.
    }
    Compute the joint loss $\mathcal{L}^t$ using $\mathbf{x}^t$ with Eq. \eqref{eq:finalloss}\;
    \If {$\mathbf{x}_i^t \in L$}{
    	Update class center $\mathbf{c}^t$ with Eq. \ref{eq:center}: $\mathbf{c}^t = \mathbf{c}^{t-1} + \alpha \Delta \mathbf{c}^t$.
    }
    Backward propagation\;
    Update the parameter set $\theta^t$\;
    }
\nl {\bf return:} $\hat{\theta} = \theta^T$
 \caption{{\bf The semi-supervised feature learning with proposed pseudo-labeling approach}}
 \label{alg:overall}
 \end{algorithm}

%% file: tex/center.tex
\begin{figure}
\centering
\includegraphics[width=0.5\textwidth]{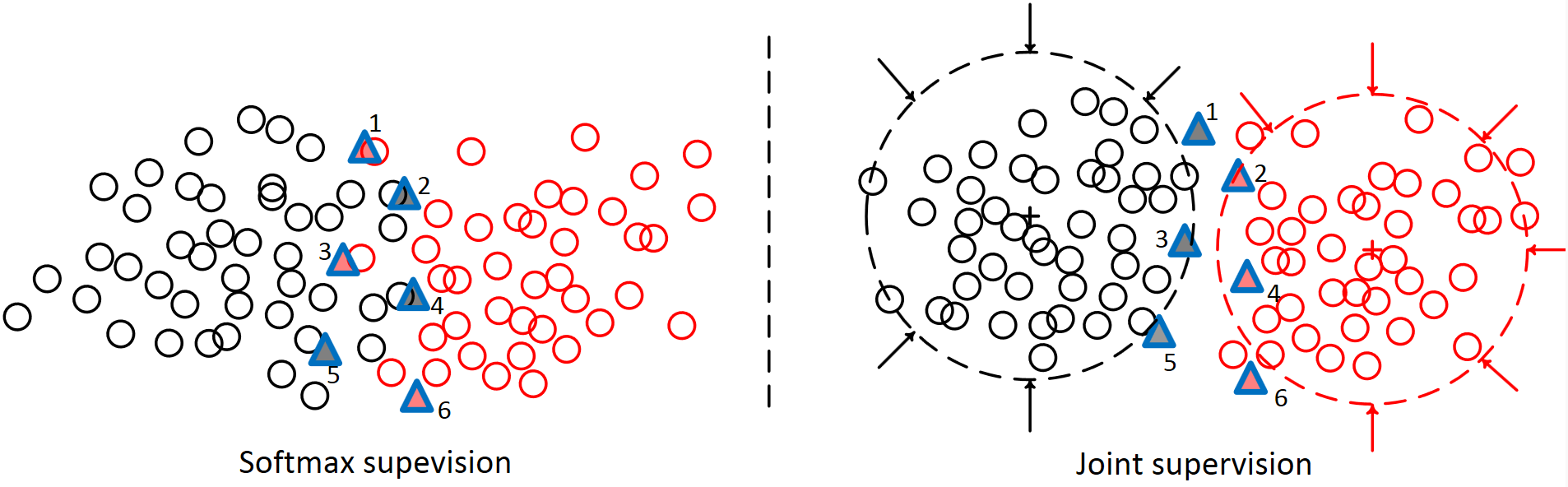}
\caption{This figure illustrates the distributions of different samples in the feature space before and after the center loss is imposed. Circles with different colors (e.g. red and black) stand for feature representations of {\it real (labeled)} samples form different categories. Blue triangles represent GAN generated {\it fake (unlabeled)} data. Generated data samples are denoted 1-6 from top to bottom. The filled color of each triangle denotes class label predicted under each case. Cross in each dashed circle denotes the center of that class. (Best viewed in color.)}
\label{fig:center}
\end{figure}

%% file: tex/comp.tex
\begin{table}
\centering
\caption{Comparison with closely related work from different aspects.}
\scalebox{0.9}{
\renewcommand\arraystretch{1.1}
\begin{tabular}{@{}l|c|c|c|c@{}}
\hline
\multicolumn{1}{l|}{Method} & \multicolumn{1}{l|}{\begin{tabular}[c]{@{}c@{}}Label\\ Assignment\end{tabular}} & \multicolumn{1}{l|}{\begin{tabular}[c]{@{}c@{}}Label\\ Distribution\end{tabular}} & \multicolumn{1}{l|}{\begin{tabular}[c]{@{}c@{}}Label\\ Source\end{tabular}} & \multicolumn{1}{l}{\begin{tabular}[c]{@{}c@{}}Contributions on\\ Pre-defined Classes\end{tabular}} \\ \hline
All-in-one\cite{odena2016semi} & static & one-hot & Manual & - \\
One-hot\cite{lee2013pseudo} & dynamic & one-hot & probability & - \\
LSRO\cite{Zheng2017Unlabeled} & static & distributed & Manual & same \\
MpRL\cite{huang2018multi} & dynamic & distributed & probability & different \\\hline
FAPL-o & dynamic & one-hot & similarity & - \\
FAPL-d & dynamic & distributed & similarity & different \\ \hline
\end{tabular}}
\label{tab:comp}
\end{table}

%% file: tex/effect.tex
\begin{table}[]
\centering
\caption{Comparison with current state-of-art pseudo-labeling methods for person re-identification, namely, LSRO and MdRL on Market-1501 and DukeMTMC-reID datasets. 24,000 generated data are incorporated for training. }
\label{tab:effect}
\begin{tabular}{l|cc|cc}
\hline
\multirow{2}{*}{Methods} & \multicolumn{2}{c|}{Market-1501} & \multicolumn{2}{c}{DukeMTMC-reID} \\ \cline{2-5} 
 & rank-1 & mAP & rank-1 & mAP \\ \hline
Baseline & 72.74 & 50.99 & 65.22 & 44.99 \\ \hline
LSRO\cite{Zheng2017Unlabeled} & 78.21 & 56.33 & 67.68 & 47.13 \\ 
dMpRL-II\cite{huang2018multi} & 80.37 & 58.59 & 68.24 & 48.58 \\ \hline
FAPL-o (Ours) & 82.04 & 61.26 & 70.92 & 51.99 \\ 
FAPL-d (Ours) & \textbf{83.43} & \textbf{63.23} & \textbf{71.90} & \textbf{52.25} \\ \hline
\end{tabular}
\end{table}

%% file: tex/gannumbers.tex
\begin{table*}[htp]
\caption{Rank-1 accuracy (\%) and mAP (\%) on Market-1501 dataset with varying numbers of unlabeled training data. Best results amongst approaches are in bold whilst best results for each method with different number of unlabeled data samples are underlined. }
\vspace{-0cm}
\begin{center}
\scalebox{0.9}{
\renewcommand\arraystretch{1.2}
\begin{tabular}{c|cc|cc|cc|cc|cc|cc|cc|cc}
\hline
\multirow{2}{*}{\# GAN images} & \multicolumn{2}{c|}{All-in-one \cite{odena2016semi,salimans2016improved}} & \multicolumn{2}{c|}{One-hot \cite{lee2013pseudo}} & \multicolumn{2}{c|}{LSRO \cite{Zheng2017Unlabeled}} & \multicolumn{2}{c|}{sMpRL \cite{huang2018multi}} & \multicolumn{2}{c|}{dMpRL-I \cite{huang2018multi}} & \multicolumn{2}{c|}{dMpRL-II \cite{huang2018multi}} & \multicolumn{2}{c|}{CPL-o(Ours)} & \multicolumn{2}{c}{CPL-d(Ours)} \\ \cline{2-17} 
 & rank-1 & mAP & rank-1 & mAP & rank-1 & mAP & rank-1 & mAP & rank-1 & mAP & rank-1 & mAP & rank-1 & mAP & rank-1 & mAP \\ \hline
0 (baseline) & 72.74 & 50.99 & 72.74 & 50.99 & 72.74 & 50.99 & 72.74 & 50.99 & 72.74 & 50.99 & 72.74 & 50.99 & 72.74 & 50.99 & 72.74 & 50.99 \\ \hline
12000 & 76.96 & 55.68 & 76.52 & 55.69 & 77.17 & 55.22 & 77.73 & 55.27 & 77.88 & 55.84 & 79.22 & 58.14 & 81.38 & 60.31 & \bbest{83.28} & \bbest{61.68} \\
18000 & \best{77.40} & 55.59 & 77.95 & 55.04 & 76.96 & 55.28 & 77.73 & 55.05 & 78.36 & 56.21 & 79.81 & 58.31 & 82.10 & \best{62.31} & \bbest{83.16} & \bbest{62.38} \\
24000 & 77.21 & 56.07 & 77.62 & \best{56.90} & \best{78.21} & \best{56.33} & \best{78.85} & 55.59 & 77.79 & 56.10 & \best{80.37} & \best{58.59} & 82.04 & 61.26 & \bbest{\best{83.43}} & \bbest{\best{63.23}} \\
30000 & 77.17 & \best{56.19} & \best{77.95} & 56.54 & 77.46 & 55.40 & 77.82 & \best{55.76} & 78.65 & 57.15 & 79.16 & 57.69 & 82.10 & 61.42 & \bbest{83.02} & \bbest{62.41} \\
36000 & 75.92 & 55.24 & 77.42 & 56.38 & 77.91 & 55.82 & 78.32 & 55.45 & \best{78.95} & \best{57.42} & 79.90 & 57.61 & \best{82.12} & 60.70 & \bbest{82.30} & \bbest{61.92} \\ \hline
Perf. boost & 4.66 & 5.20 & 5.21 & 5.91 & 5.47 & 5.34 & 6.11 & 4.77 & 6.21 & 6.43 & 7.63 & 7.60 & 9.38 & 11.32 & \bbest{\best{10.69}} &\bbest{\best{12.24}} \\ \hline
\end{tabular}}
\end{center}
\label{tab:gannumbers}
\vspace{-0cm}
\end{table*}

%% file: tex/reduced.tex
\begin{table}[]
\centering
\caption{Results on Market1501 dataset with reduced labeled data subsets. This is trained with distributed pseudo-labels. Best performance for each reduction case is shown in bold. }
\label{tab:reduced}
\scalebox{0.95}{
\begin{tabular}{c|cc|cc|cc}
\hline
\multirow{2}{*}{number of images} & \multicolumn{2}{c|}{All} & \multicolumn{2}{c|}{half} & \multicolumn{2}{c}{third} \\ \cline{2-7} 
 & rank-1 & mAP & rank-1 & mAP & rank-1 & mAP \\ \hline
0(baseline) & 72.74 & 50.99 & 66.98 & 43.71 & 57.45 & 33.22 \\ \hline
12000 & 83.28 & 61.68 & 76.81 & 54.25 & 66.39 & 42.87 \\ 
18000 & 82.16 & 61.68 & 77.02 & 54.48 & 65.23 & 41.22 \\ 
24000 & \textbf{83.43} & 62.23 & 77.46 & 55.26 & 66.48 & 42.04 \\ 
30000 & 83.02 & \textbf{62.41} & \textbf{78.59} & \textbf{56.50} & 66.69 & 43.50 \\ 
36000 & 82.30 & 61.92 & 78.15 & 56.30 & \textbf{67.87} & \textbf{43.54} \\ \hline
\end{tabular}}
\end{table}

%% file: tex/param.tex
\begin{table}[]
\centering
\caption{Sensitivity analysis for the trade-off parameter $\lambda$ balancing the contributions of classification loss and center loss. Experiments are performed on Market-1501 with 24,000 }
\label{tab:param}
\begin{tabular}{c|cc|cc}
\hline
\multirow{2}{*}{lambda} & \multicolumn{2}{c|}{one-hot} & \multicolumn{2}{c}{Distributed} \\ \cline{2-5} 
 & rank-1 & mAP & rank-1 & mAP \\ \hline
0.001 & 80.82 & 60.42 & 81.05 & 61.18 \\ \hline
0.0001 & \bbest{82.04} & \bbest{61.26} & \bbest{83.43} & \bbest{63.23} \\ \hline
0.00001 & 81.18 & 59.42 & 82.31 & 62.31 \\ \hline
\end{tabular}
\end{table}

%% file: tex/gans.tex
\begin{figure}[ht]

\subfigure[Real]{
\includegraphics[width=\columnwidth]{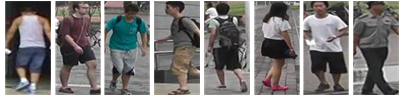}
\label{fig:real}
}\vspace{-0.2cm}
\quad \quad

\subfigure[DCGAN]{
\includegraphics[width=\columnwidth]{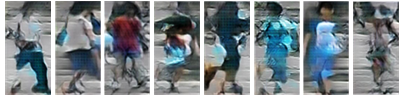}
\label{fig:dcgan}
}\vspace{-0.2cm}
\quad\quad
\subfigure[IWGAN]{
\includegraphics[width=\columnwidth]{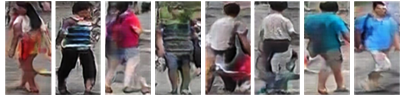}
\label{fig:iwgan}
\vspace{-0.2cm}
}
\caption{Examples of real and GAN generated images on Market-1501. (a) The top row shows the real pedestrian samples from training set. (b) The middle row shows generated images from DCGAN, the visual quality is relatively low from the perspectively of a human viewer, but they can help regularize the model. (c) The bottom row shows the generated images from IWGAN. Better visual qualities in human body shape and clothing can be observed. }
\label{fig:gans}
\end{figure}

%% file: tex/iwgan.tex
\begin{table*}[]
\centering

\caption{Rank-1 accuracy (\%) and mAP (\%) results on Market-1501 and DukeMTMC-reID datasets when two GAN models, DCGAN and IWGAN, are adopted for unlabeled image generation. For each dataset, best performance for one-hot and distributed schemes are underlined and in bold, respectively. 
}
\scalebox{0.9}{
\renewcommand\arraystretch{1.2}
\begin{tabular}{c|cc|cc|cc|cc|cc|cc|cc|cc}
\hline
\multirow{4}{*}{\# GAN images} & \multicolumn{8}{c|}{Market} & \multicolumn{8}{c}{DukeMTMC-reID} \\ \cline{2-17} 
 & \multicolumn{4}{c|}{DCGAN \cite{radford2015unsupervised}} & \multicolumn{4}{c|}{IWGAN\cite{gulrajani2017improved}} & \multicolumn{4}{c|}{DCGAN\cite{radford2015unsupervised}} & \multicolumn{4}{c}{IWGAN\cite{gulrajani2017improved}} \\ \cline{2-17} 
 & \multicolumn{2}{c|}{One-hot} & \multicolumn{2}{c|}{Distributed} & \multicolumn{2}{c|}{One-hot} & \multicolumn{2}{c|}{Distributed} & \multicolumn{2}{c|}{One-hot} & \multicolumn{2}{c|}{Distributed} & \multicolumn{2}{c|}{One-hot} & \multicolumn{2}{c}{Distributed} \\ \cline{2-17} 
  & rank-1 & mAP & rank-1 & mAP & rank-1 & mAP & rank-1 & mAP & rank-1 & mAP & rank-1 & mAP & rank-1 & mAP & rank-1 & mAP \\ \hline
12000 & 81.38 & 60.31 & \bbest{83.28} & 61.68 & \best{82.84} & \best{62.87} & 83.05 & \bbest{62.60} & 71.57 & 52.68 & 71.68 & 52.83 & \best{72.21} & \best{53.23} & \bbest{73.16} & \bbest{54.96} \\ 
18000 & 82.10 & 62.31 & 82.16 & 61.18 & \best{82.17} & \best{62.50} & \bbest{82.89} & \bbest{62.26} & 70.38 & 51.87 & 71.32 & 52.88 & \best{71.98} & \best{53.34} & \bbest{72.60} & \bbest{53.97} \\ 
24000 & 82.04 & 61.26 & 83.43 & 63.23 & \best{82.42} & \best{61.58} & \bbest{83.84} & \bbest{63.41} & 70.92 & 51.99 & 71.90 & 52.25 & \best{71.72} & \best{53.01} & \bbest{72.35} & \bbest{54.00} \\ 
30000 & 82.10 & 61.42 & 83.02 & 62.41 & \best{82.66} & \best{61.62} & \bbest{83.58} & \bbest{63.78} & 70.47 & 52.54 & 72.38 & 53.71 & \best{72.48} & \best{53.35} & \bbest{73.44} & \bbest{54.71} \\ 
36000 & 82.12 & 60.70 & 82.30 & 61.92 & \best{82.51} & \best{61.29} & \bbest{82.78} & \bbest{63.43} & 71.23 & 52.18 & 72.40 & 53.73 & \best{72.40} & \best{52.76} & \bbest{72.85} & \bbest{53.85}\\ \hline
\end{tabular}}
\label{tab:iwgan}
\end{table*}

%% file: tex/market.tex
\begin{table}[]
\caption{Comparison with state-of-art methods on market-1501 dataset. Best and second best results are denoted as bold and underlined text, respectively.}
\begin{center}
\begin{tabular}{l|cc}
\hline
\multirow{2}{*}{Method} & \multicolumn{2}{c}{Market 1501} \\ \cline{2-3} 
 & rank-1 & mAP \\ \hline
Gate-reID \venue{(ECCV'16)} \cite{varior2016gated} & 65.88 & 39.55\\
SCSP \venue{(CVPR'16)} \cite{chen2016similarity}& 51.90 & 26.35 \\
DNS \venue{(CVPR'16)} \cite{zhang2016learning} & 61.02 & 35.68 \\
ResNet+OIM \venue{(CVPR'17)} \cite{xiao2017joint} & 82.10 & -  \\
Latent Parts \venue{(CVPR'17)} \cite{li2017learning}& 80.31 & 57.53 \\
P2S \venue{(CVPR'17)} \cite{zhou2017point} & 70.72 & 44.27\\
Consistent-Aware \venue{(CVPR'17)} \cite{lin2017consistent}  & 80.90 & 55.60\\
Spindle \venue{(CVPR'17)} \cite{zhao2017spindle} & 76.90 & - \\
SSM \venue{(CVPR'17)} \cite{bai2017scalable}  & 82.21 & \best{68.80} \\
JLML \venue{(IJCAI'17)} \cite{li2017person}  & \best{85.10} & 65.50 \\
SVDNet \venue{(ICCV'17)} \cite{sun2017svdnet}  & 82.30 & 62.10\\
Part Aligned \venue{(ICCV'17)} \cite{zhao2017deeply}  & 81.00 & 63.40 \\
PDC \venue{(ICCV'17)} \cite{su2017pose}  & 84.14 & 63.41 \\
LSRO \venue{(ICCV'17)} \cite{Zheng2017Unlabeled} & 78.06 & 56.23 \\
dMpRL-II \venue{(Arxiv'18)} \cite{huang2018multi} &80.37 & 58.59 \\
\hline
Baseline & 72.74 & 50.99\\
Ours-o+DCGAN & 82.10 & 62.31 \\
Ours-d+DCGAN & 83.43 & 63.23 \\\hline
Ours-o+IWGAN & 82.66 & 61.62 \\
Ours-d+IWGAN & 83.58 & 63.78 \\
Ours-d+IWGAN+re-rank & \bbest{86.07} & \bbest{77.64} \\ \hline
\end{tabular}
\end{center}
\label{tab:market}
\end{table}

%% file: tex/duke.tex
\begin{table}[h]
\caption{Comparison of state-of-art approaches on the DukeMTMC-reID dataset. Rank-1 accuracy (\%) and mAP (\%) are reported.}
\begin{center}
\begin{tabular}{l|cc}
\hline
\multirow{2}{*}{Method} & \multicolumn{2}{c}{DukeMTMC-reID} \\ \cline{2-3} 
 & rank-1 & mAP \\ \hline
BOW+kissme \venue{(ICCV'15)}\cite{zheng2015scalable} & 25.13 & 12.17 \\
LOMO+XQDA \venue{(CVPR'15)}\cite{liao2015person} & 30.75 & 17.04 \\
LSRO \venue{(ICCV'17)}\cite{Zheng2017Unlabeled} & 67.68 & 47.13 \\
dMpRL \venue{(Arxiv'18)}\cite{huang2018multi} & 68.24 & 48.58 \\
Verif + Identif \venue{(TOMM'17)}\cite{zheng2017discriminatively} & 68.90 & 49.30 \\
APR \venue{(Arxiv'17)}\cite{lin2017improving} & 70.69 & 51.88 \\
ACRN \venue{(CVPRW'17)}\cite{schumann2017person} & 72.58 & 51.96 \\
PAN \venue{(Arxiv'17)}\cite{zheng2017pedestrian} & 71.59 & 51.51 \\
FMN \venue{(Arxiv'17)}\cite{ding2017let} & 74.51 & 56.88 \\
Bilinear Coding \venue{(Arxiv'18)} \cite{zhou2018weighted} & 76.20 & 56.90 \\
SVDNet \venue{(ICCV'17)}\cite{sun2017svdnet} & 76.70 & 56.80 \\
DPFL \venue{(ICCVW'17)}\cite{chen2018person} & \bbest{79.20} & \best{60.60} \\ \hline
Baseline & 65.22 & 44.99 \\
Ours-o+DCGAN & 71.57 & 52.68 \\
Ours-d+DCGAN & 72.38 & 53.71 \\\hline
Ours-o+IWGAN & 72.40 & 52.76 \\
Ours-d+IWGAN & 72.85 & 53.85 \\
Ours-d+IWGAN+re-rank & \best{79.04} & \bbest{70.74} \\ \hline
\end{tabular}
\end{center}
\label{tab:duke}
\end{table}